%% file: digitalTwin.tex
\documentclass[a4paper,10pt,twocolumn]{article}
\pdfoutput=1
\usepackage[utf8]{inputenc}
\usepackage[german,english]{babel}
\usepackage[T1]{fontenc}

\usepackage{ragged2e}
\usepackage{eurosym}
\usepackage{url}
\usepackage{pdfpages}
\usepackage{todonotes}
\usepackage{figsize}
\usepackage{amssymb} 
\usepackage{amsmath}
\usepackage{amsthm}
\usepackage{xspace}
\usepackage{graphicx}
\usepackage{stfloats}
\usepackage{float}
\usepackage{algorithmic}
\usepackage{booktabs}
\usepackage{multirow}

\DeclareGraphicsExtensions{.pdf,.ai,.jpg,.png}

\usepackage{color, colortbl}

\usepackage{xcolor}
\definecolor{Gray}{gray}{0.9}

\newcommand{\bsfigure}[4][tb]{%
  \begin{figure}[#1]
    \centering
    \vspace*{-10px}
    \includegraphics[scale=#2]{#3}
    \vglue 0ex plus 0.5ex minus 0.5ex
    {\caption{\small #4.}\label{#3}}%
  \end{figure}}

\newcommand{\nat}{\ensuremath{\mathbf{N} }}

\newcommand{\real}{\ensuremath{\mathbf{R} }}

\usepackage{enumitem}
\setlist[enumerate]{topsep=4pt,itemsep=4pt}

\newcommand{\Ni}{(1)~}
\newcommand{\Nii}{(2)~}

\newcommand{\dt}{DigitalTwin~}

\begin{document}

\setlength{\textwidth}{\textwidth+1.0ex}

\title{\vspace*{-12ex}The DigitalTwin from an Artificial Intelligence Perspective}
\date{\vspace*{-5ex}}

\author{Oliver Niggemann$^1$, oliver.niggemann@hsu-hh.de, 
\\ 
Alexander Diedrich$^2$, alexander.diedrich@iosb-ina.fraunhofer.de,
  \\ 
Christian Kühnert$^2$, , christian.kuehnert@iosb.fraunhofer.de,
\\ 
Erik Pfannstiel$^1$, erik.pfannstiel@hsu-hh.de,
\\ 
Joshua Schraven$^1$, joshua.schraven@hsu-hh.de\\
  \\
  $^1$Institute of Automation Technology, Helmut Schmidt University,  Hamburg, Germany,  \\
  $^2$Fraunhofer IOSB, Fraunhofer Center for Machine Learning, 
Karlsruhe, Germany  
}

\maketitle

\begin{abstract}
Services for Cyber-Physical Systems based on Artificial Intelligence and Machine Learning require a virtual representation of the physical. To reduce modeling efforts and to synchronize results, for each system, a common and unique virtual representation used by all services during the whole system life-cycle is needed---i.e. a DigitalTwin. In this paper such a DigitalTwin, namely the AI reference model AITwin, is defined. This reference model is verified by using a running example from process industry and by analyzing the work done in recent projects.
\end{abstract}

\input{intro.tex}

\input{SOTA.tex}

\input{reqs.tex}

\input{timing.tex}

\input{extrapolate.tex}

\input{causalities.tex}

\input{aitwin.tex}

\input{empiric2.tex}

\input{contribution.tex} 


\bibliographystyle{abbrv}
\bibliography{dtbib}

\end{document}

%% file: intro.tex
\section{Introduction} \label{intro}
DigitalTwins are a key concept for Cyber-Physical-Systems (CPS): By maintaining a collection of relevant information about a physical entity, a digital shadow is created which can be used for tasks such as monitoring, diagnosis or optimization. Most publications about the \dt focus on engineering- and process-oriented aspects such as a continuous enrichment of the twin during its life-cycle \cite{TAO2019653}, on simulation scenarios \cite{rosena:2019a, systems7010007} or on modeling questions such as optimal meta-levels \cite{stark:2019a}, hierarchies \cite{TAO2019653} or engineering chains \cite{systems7010007}. Even the few publications written with a clear focus on Artificial Intelligence (AI) and Machine Learning (ML) fail to connect the content and capabilities of the \dt with AI/ML methods. A review of the related work confirming this impression is given in section \ref{subsec:related_work}. Under the bottom-line, the \dt is seen as an information hole into which all available information is poured---hoping for a benefit by AI/ML applications at some later point in the life-cycle. 

On the other hand, AI/ML methods have always used models of the environment and domain knowledge. So the \dt concept and AI/ML should be compatible. But two main contradictions remain: First, AI/ML are very heterogeneous, and each AI/ML method comes with a specialized model formalism to capture relevant aspects of the environment and the application domain. Hence, the question is how a \dt can provide the correct model to each AI/ML method. The second contradiction is that  AI/ML requires explicit, i.e. by an algorithm processable knowledge, since compiled knowledge in form of simulation libraries, raw data or executables does not help. But most publications refer to these kind of information.

In this paper we follow the key idea in Edward E. Lee's seminal paper \cite{Lee:2010:CF:1837274.1837462} in which he claims that ``The intellectual heart of CPS is in studying the joint dynamics of physical processes, software, and networks.'', i.e. from an AI/ML perspective the main task of a \dt is to use models to predict the behavior of the corresponding CPS subsystem, comprising cyber and physical parts. Here we add an additional constraint: The prediction knowledge must be explicit and processable by an AI/ML algorithm. 

So the paper addresses the following research questions (RQs):

\noindent \textit{RQ 1:} Can we develop, as shown in figure \ref{RQ1.pdf}, a common API of a DigitalTwin which is suitable for a set of heterogeneous AI algorithms? I.e. here in this paper neither the AI algorithms nor the internal structure of the DigitalTwin is addressed but only a common DigitalTwin API.

\bsfigure[htb!]{0.32}{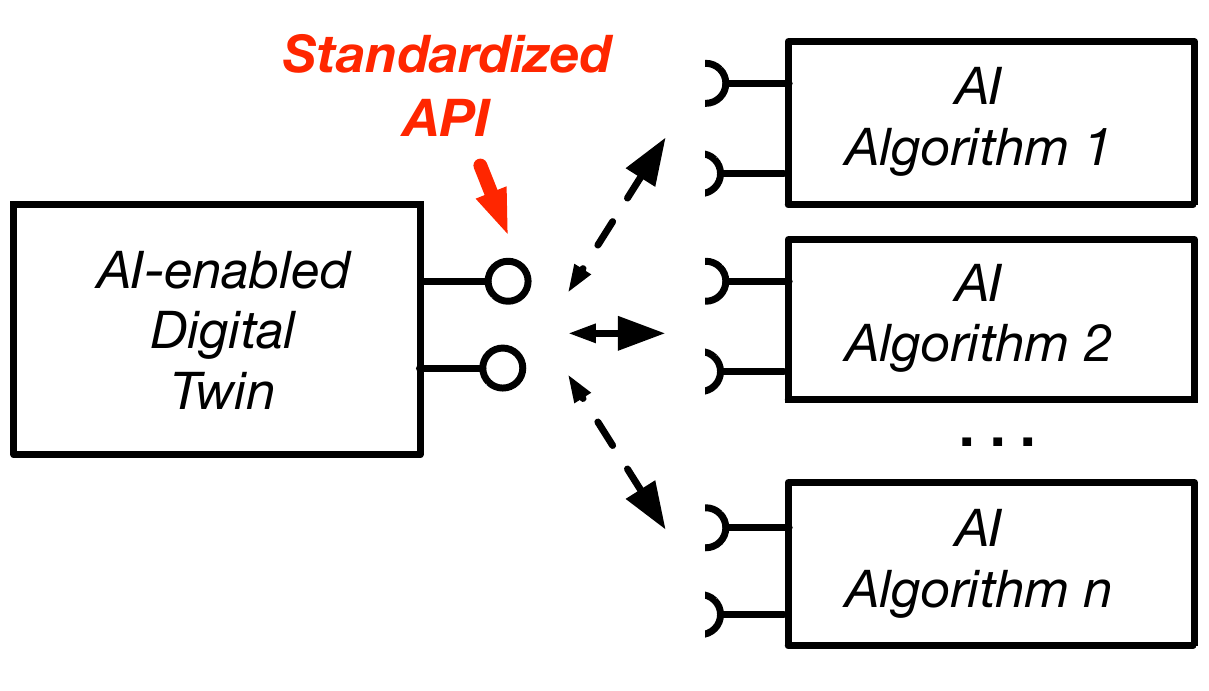}{Research Questions 1: Common AI-enabled API for DigitalTwins}

\noindent \textit{RQ 2:} Can AI algorithms use a common information base, stored in the DigitalTwin? Can AI algorithms even exchange information that way? 

\noindent \textit{RQ 3:} Can such a DigitalTwin be used for both machine learning algorithms (i.e. sub-symbolic AI) and symbolic AI algorithms? Bridging sub-symbolic and symbolic AI is currently a major topic in the AI community \cite{Lample2020DeepLF,Besold2017NeuralSymbolicLA}.

In section \ref{reqs}, an analysis of the requirements of AI/ML methods to the corresponding models is made and in section \ref{section3} the main solution approach is presented. As a result, a \dt AI reference model, called the \textbf{AITwin}, will be derived. If such a reference model is used by several AI/ML methods, multiple advantages arise: First of all, the results of one method could be used by another. Second, redundant information used by different methods are avoided, easing the modeling efforts and preventing contradicting behavior. Case studies can be found in section \ref{empirical}, they are used to evaluate the solution approach. A conclusion is given in section \ref{contribution}.

%% file: SOTA.tex
\section{Related work} \label{subsec:related_work}

As described in the introduction, the DigitalTwin with its enabling technologies plays a decisive role for the success in several areas and is an ongoing field of research. One of the most comprehensive reviews for the current-state-of-the-art has been made by Fuller \cite{fuller2019digital} covering the areas manufacturing, smarts cities and healthcare. As one of his conclusions he states that there is still the lack of \dt reference models, which leads to discrepancies between similar projects and hence slows done the progress of this technology. Proposals for \dt reference models have already been made in \cite{Bevilacqua:2020} with the focus on risk control and prevention on process plants, in \cite{lu:2018} for smart-manufacturing and in \cite{alam:2017} for cloud-based CPS. 

Recent publications, taking into account DigitalTwins and AI/ML are given by Min \cite{min2019machine} focussing on production optimization in petrochemical industry or \cite{droder:2018} who proposes a machine-learning enhanced \dt for Human-Robot-Collaboration. Still, most of the research is dealing with DigitalTwins from a modeling perspective. For example \cite{luo:2019} develops the \dt of a CNC-machine, in \cite{tuegel:2011} a aircraft has been reengineered and \cite{rasheed:2020} explains the challenges and enablers for \dt in modeling.

Hence, current research combining AI with DigitalTwins results always in individual solutions for specific application fields. The herein presented work tries to close this gap by proposing the \dt AI reference model AITwin.

%% file: reqs.tex
\section{AI/ML Methods and the \dt} \label{reqs}

Here we categorize AI/ML methods and derive requirements to the DigitalTwin. The number of used AI/ML methods are legion and so are the formalisms used to model domain knowledge and environment models. Still for simplicity sake, we focus on those methods mainly used for CPS and categorize these AI/ML methods into three main categories shown on the left hand side of figure \ref{levels.pdf}. To illustrate these applications, a running example is used in this paper (figure \ref{ZeichnungDemoCase.pdf}):
\bsfigure[htb!]{0.32}{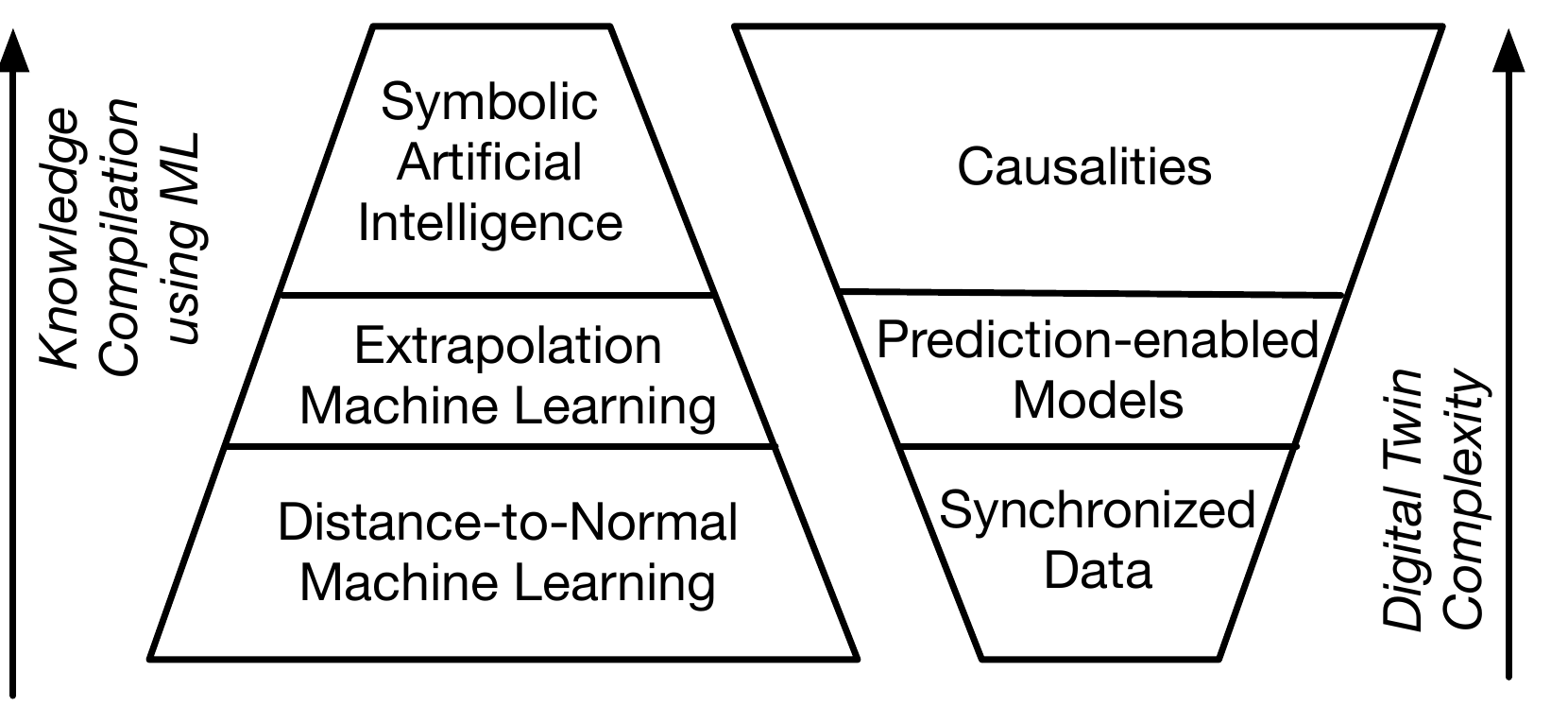}{Levels of AI/ML and the matching complexity of digital twins}

The system consists of four water tanks $t$, seven electric valves $p$ with integrated flow sensors, an unlimited water source and an unlimited water sink (not shown). Valve $v_0$ controls water from the unlimited water source, for example the public water mains, into tank $t_0$. From there, three pipes with an equal diameter divide the water flow. Finally, valve $v_6$ drains tank $t_3$ into the unlimited water sink, for example a river or a processing facility.
\bsfigure[htb!]{0.27}{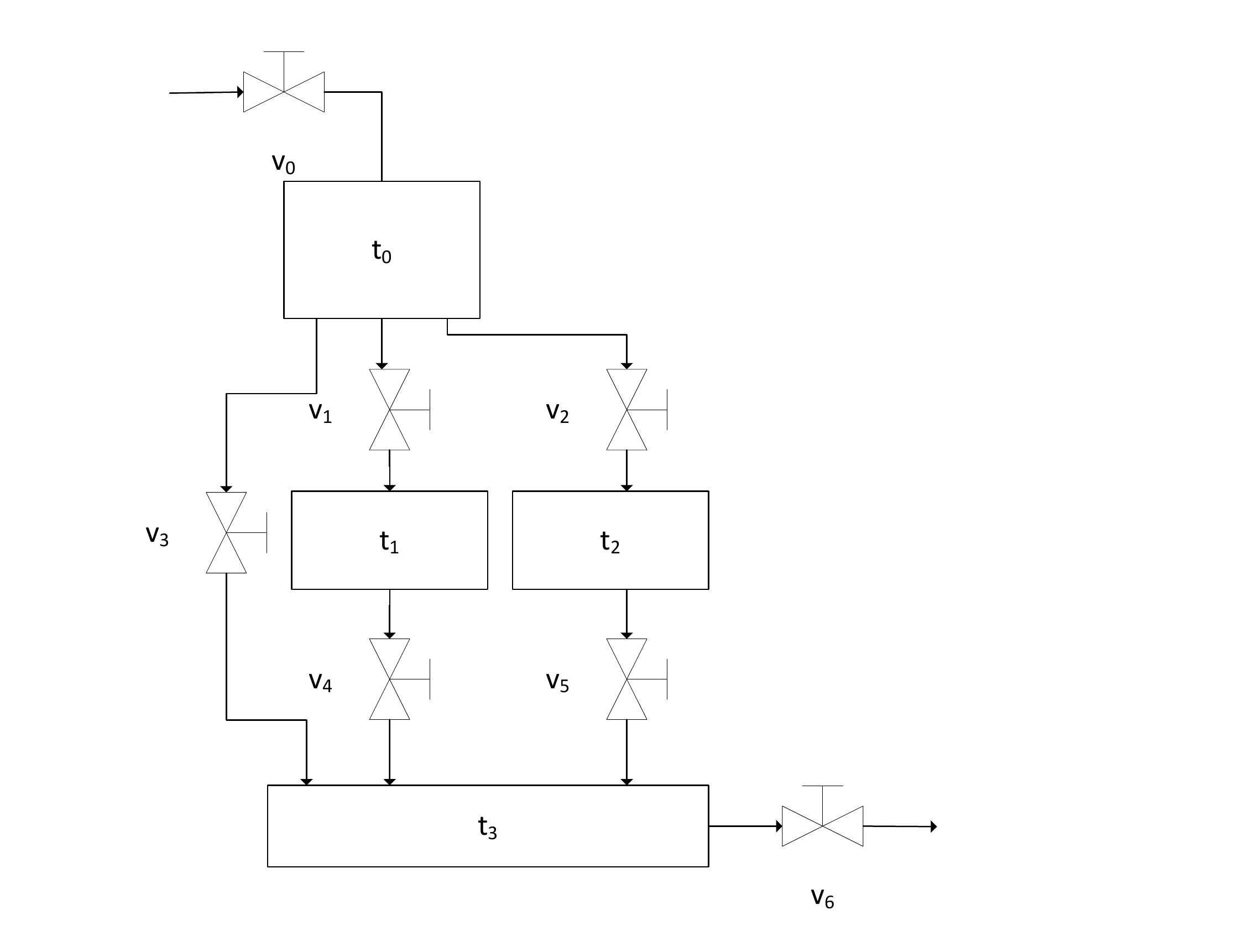}{The running example in the form of a four tank model}

\noindent \textbf{Distance-to-Normal Machine Learning:}  Anomaly detection is a typical example for this approach: The machine learner memorizes historical data points and computes a distance measure between a current data point (e.g. production operation point) and memorized points. Often, the historical data points are interpreted as ``normal'' and too distant data points are classified as ``anomalous''. To some small extent, interpolation is used to deal with data points close to previous data points. To implement ``Distance-to-Normal Machine Learning'', the digital twin (see right hand side of figure \ref{levels.pdf}) must comprise a common timing model to synchronize all data.
	
	In our running example, the vector of all flows can be used to detect unusual flows, i.e. anomalies.
	
\noindent \textbf{Extrapolation Machine Learning:} Using a \dt in autonomous systems for optimization and self-configuration means using it in a closed control loop, e.g. for model-predictive control tasks. I.e. the \dt must predict output values such as resource consumption, positions, etc. for new input values. Please notice that this extrapolation capability means to predict values, in time or distance, far away from observed data points. In terms of ML this is especially a problem for technical systems where often only few individual data points are available which are also clustered in only a few regions.

	In our running example, the vector of all flows can be prognosed (i.e.extrapolated) for new valve positions. This can be used to optimize the system configuration.

\noindent \textbf{Symbolic Artificial Intelligence:} Many AI tasks such as diagnosis, reconfiguration or planning require causal information, i.e. symbolic AI models: Diagnosis uses root cause-symptom relations (``if the tube leaks, the tank fills slowly") and planning needs information about process steps (``Modul 1 takes a board and drills a  hole''). Such causalities always describe effects caused by causes where both effects and causes are symbolic predicates in some formal logic. So the digital twin (see right hand side of figure \ref{levels.pdf}) must comprise such logic-based models.

	In our running example, the effect of switching a valve such as an overflowing container can be modeled, e.g. to identify root causes of errors.

%% file: timing.tex
\section{AI Reference Model}\label{section3}
In this section, the three research questions from section \ref{intro} are discussed and, as a result, the \dt AI reference model, \textbf{AITwin}, is derived.

\subsection{Timing Model} \label{timing}
While time is the essential feature of any physical process (the ''P'' in CPS), software (the ''C'' in CPS) does not come with any build-in concept of time. In software parallel events may occur at various points in time and time spans between commands are not defined but vary according to process behavior. So first, a common time model of the \dt must be established. Following an idea from \cite{Lee:2010:CF:1837274.1837462}, one may differentiate between two general solutions for this: 

\noindent \emph{Cyberizing the Physical:} Normally such approaches define signal values $x$ as $x(t,k)$ where $t \in \real$ is the physical time and $k\in \nat$ the number of an ordered sequence of events/instructions in the software. This is called a superdense model of time. 

\noindent \emph{Physicalizing the Cyber:} New means for the time synchronization in distributed systems, e.g. the Precision Time Protocol \cite{ieee1588}, allow for a consistent concept of physical time especially in controllers. Hence, software now may refer to precise points in time and time spans.\newline
''Cyberizing the Physical'' has the drawback that most AI algorithms need to be adapted, since they require a clear time model and concept of causality. Therefore, the usage of ''Physicalizing the Cyber'' approach should be prefered, i.e. all information $\mathbf{x}(t)$ in the \dt refers to a unique point in time $t$. This requires from the underlying technical systems time synchronization between all devices and an implemented concept of simultaneity. 

So on this level only two interfaces are needed for the \dt AI reference model (AITwin): Function \textit{getData(i,t)} returns the i'th signal $x \in \real$ at a specific point in time $t$ and function \textit{getData(t)} returns all signals $\mathbf{x} \in \mathcal{X}$ at a specific point in time $t$. We refer with $\mathcal{X}$ to the set of all possible values $\mathbf{x}(t)$. 

\bsfigure[htb!]{0.42}{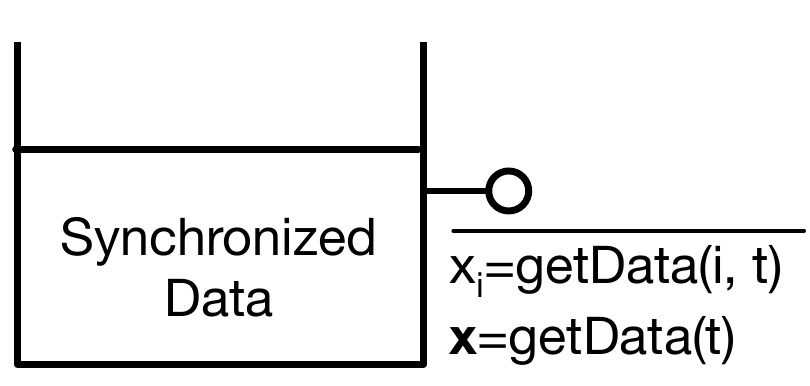}{Interfaces for the first level of the \dt AI reference model AITwin}
Please note that the vector $\mathbf{x}(t)$ captures the complete observable state of the system, e.g. comprising sensor/actuator signals, observable product/object locations and states, configuration parameters and so on.

%% file: extrapolate.tex
\subsection{Extrapolatable Models} \label{extrapolate}
While superficially much data seems to be generated by Cyber Physical Systems, a different picture can be seen when ML is applied: Data intensive ML such as deep learning often fails or delivers unusable results in production settings while in other domains it show break-thru performance. This is mainly due to the following two reasons: \Ni Most data comprises repetitive patterns --mirroring the repetitive structure of typical production processes, while the pure amount of data is sufficient. \Nii The data hardly comprises information about failures or other problems. 

Therefore, learned models often only have limited extrapolation capabilities--which is essential for CPS. Furthermore, even if learned models such as neural networks predict output values for new input values, the prediction quality quickly decreases for values far away from observed data points in time and distance. Hence, the AI reference model, AITwin, must not only be able to extrapolate but also to assess the quality of the extrapolation. So far the best approach is the usage of probabilistic models, e.g. probabilistic or Bayesian Neural Networks \cite{DBLP:journals/corr/abs-1801-07710} which can also applied to time series analysis \cite{eitenheuer2018jdgsw}. So all predictions should not only be implemented as a point estimator but should also compute the probability that the prediction is correct. In general, we have to differentiate between two types of analyses and predictions:

\noindent \textit{\bf Static Predictions:} For tasks such as condition-monitoring or anomaly detection, only the signal values $\mathbf{x(t)} \in \real^n$ at some point in time $t$ are used. In other words, for the analysis a static feature vector \cite{eitenheuer2019hdsh} is taken into account. Thereby, the assumption is that no information is coded in the sequence of value and all necessary information is contained in the current signal values. This assumption is true for many CPSs, even for systems which have a dynamic nature them-self. Typically three use cases are implemented here by machine learning applications:

\noindent \textit{Anomaly Detection:} For a new data point $\mathbf{x} \in \real^n$ its probability $p(\mathbf{x} | X)$ given some historical data $X$ is (at least approximately) computed.

\noindent \textit{Prediction:} Given a partial feature vector $\mathbf{x'}  \in \real^n$, i.e. a feature vector with missing entries, those entries are computed in form of a complete feature vector $\mathbf{x}$, best with their probability. The given entries then become inputs and the computed entries become outputs. Of course, if some input to output configuration is not possible, this must be signaled to the user.

\noindent \textit{Optimization:} The reader may note that the partial gradient over the signals can be created by extrapolating in the vicinity of an operation point, so no special functionality of the \dt is needed here for the optimization tasks.

\noindent \textit{\bf Dynamic Predictions:} A totally different situation arises when important information is coded in the sequence of signal values over time. Again, three use cases exist:

\noindent \textit{Anomaly Detection:} Given historical data $X= \{\mathbf{x}(t=t_0), \mathbf{x}(t=t_1), \ldots,  \mathbf{x}(t=t_k)\}$, for a time step $\Delta t$ and a new data vector $\mathbf{x}(t=t_k + \Delta t)$ and its probability is computed so that $\mathbf{x}(t=t_k + \Delta t)$ is the logical continuation of the series. 

\noindent \textit{Prediction:} Given historical data $X= \{\mathbf{x}(t=t_0), \mathbf{x}(t=t_1), \ldots,  \mathbf{x}(t=t_k)\}$ and a time step $\Delta t$ the next value $\mathbf{x}(t=t_k + \Delta t)$ is computed, best with its probability distribution. 

\noindent \textit{Optimization:} The reader may note that the temporal gradient over the signals can be created by extrapolating in the vicinity of the point in time $\Delta t$, so no special functionality of the \dt is needed here.

As mentioned earlier, for reliable predictions, the model must not only capture the behavior in the normal but also in the failure modes. For example predictions for a failure mode, like a broken drive or a blocked pipe, are helpful for diagnosis tasks. Hence, the \dt must be able to activate specific failure modes by setting some components to failure states.

Figure \ref{arch3.pdf} shows the resulting API methods with extrapolation capabilities. The static case is solved by a function \textit{extrapolateStatic($\textbf{x'}$)} where \textbf{x'} is a partially filled signal vector. The function returns the missing values $\textbf{x}$ and a corresponding vector of probabilities, both with $null$ entries if not computable. This solves all three use cases discussed above.

For the dynamic case, a function \textit{extrapolateDynamic(\textbf{X}, $\Delta$t)} where \textbf{X} is a sequence of historical signal vectors up the point in time $t$ is defined. The function returns the next signal vector $\textbf{x}(t+\Delta t)$ and a corresponding vector of probabilities, both with $null$ entries if not computable. Again, this solves the three use case discussed above.

The failure modes are set by a function \textit{setFailedComps(\textbf{C'})} where \textbf{C'} is a subset of all components. If a failure mode is set, the functions \textit{extrapolateStatic} and \textit{extrapolateDynamic} predict the behavior assuming that the components are not working correctly. Another function \textit{getComps()} returns all components.

%% file: causalities.tex
\subsection{Causalities} \label{causalities}
No definition of causality exists which is accepted by the majority of researchers. We therefore first analyze the special usage of causality models in two main AI application fields of \dt and derive from there a suitable causality definition. 

\paragraph{AI Application 1 - Consistency-based Diagnosis:} 
Diagnosis is the task of computing the error cause based on incomplete symptoms, i.e. observations.Therefore, knowledge about the causalities between root causes (i.e. failures) and symptoms must be known since failures such as sensor faults or machine degradation often lead to complex symptom patterns, e.g. alarms, error messages and plant stops. Currently users often see complex alarms and symptom patterns but fail to identify quickly the correct root cause and therefore can not repair the factory efficiently. 

Often models predict the behavior only for the normal mode. Consistency-based diagnosis \cite{diedrich2019jfgsy} is able to use so-called weak fault models, i.e. models which capture the system behavior only for the OK-mode. In most cases, it uses partial system models such as 
$OK(C_{i_1})\wedge \ldots \wedge OK(C_{i_k}) \rightarrow \left( s_1 \rightarrow s_2 \wedge \ldots \wedge s_l \right), k,l\in \nat$ where $OK(C)$ denotes a correct functioning of component $C$ and $s_i \in \{0,1\}$ is a binary/symbolic representation of the current system status, e.g. $s_i \equiv \text{''Sensor Value 42 > 10''}$ or $s_j \equiv \text{''Motor 3 is on''}$. So normally $s_i \in S$ where $S$ describes all system states. In other words, if components $C_{i_1} \ldots C_{i_k}$ work correctly, the  implication describes the normal system causality. Consistency-based Diagnosis then redraws OK-assumptions until no contradictions between predictions and observation exist anymore.  

\paragraph{AI Application 2 - Planning:} Planning is the task of computing a sequence of process steps which transforms given raw objects into an also given final object. Let $P$ be the set of all possible objects. Then a process step $q_i$ is an implication $q_i: \text{in} \rightarrow \text{out}$, which takes $u$ input objects $\text{in} \in P^u$ and transforms them into $v$ output objects $\text{out}\in P^v$. We denote the set of all process steps as $Q$. Please note that parameters of process steps can be implemented by defining several individual process steps.

Based on these application scenarios, a common causality model for the AI reference model \textbf{AITwin} is derived. Figure \ref{causal1.pdf} shows two types of causalities, namely System State Causalitites (handling system changes) and Product State Causalities (handling product changes) are defined :
\bsfigure[htb!]{0.34}{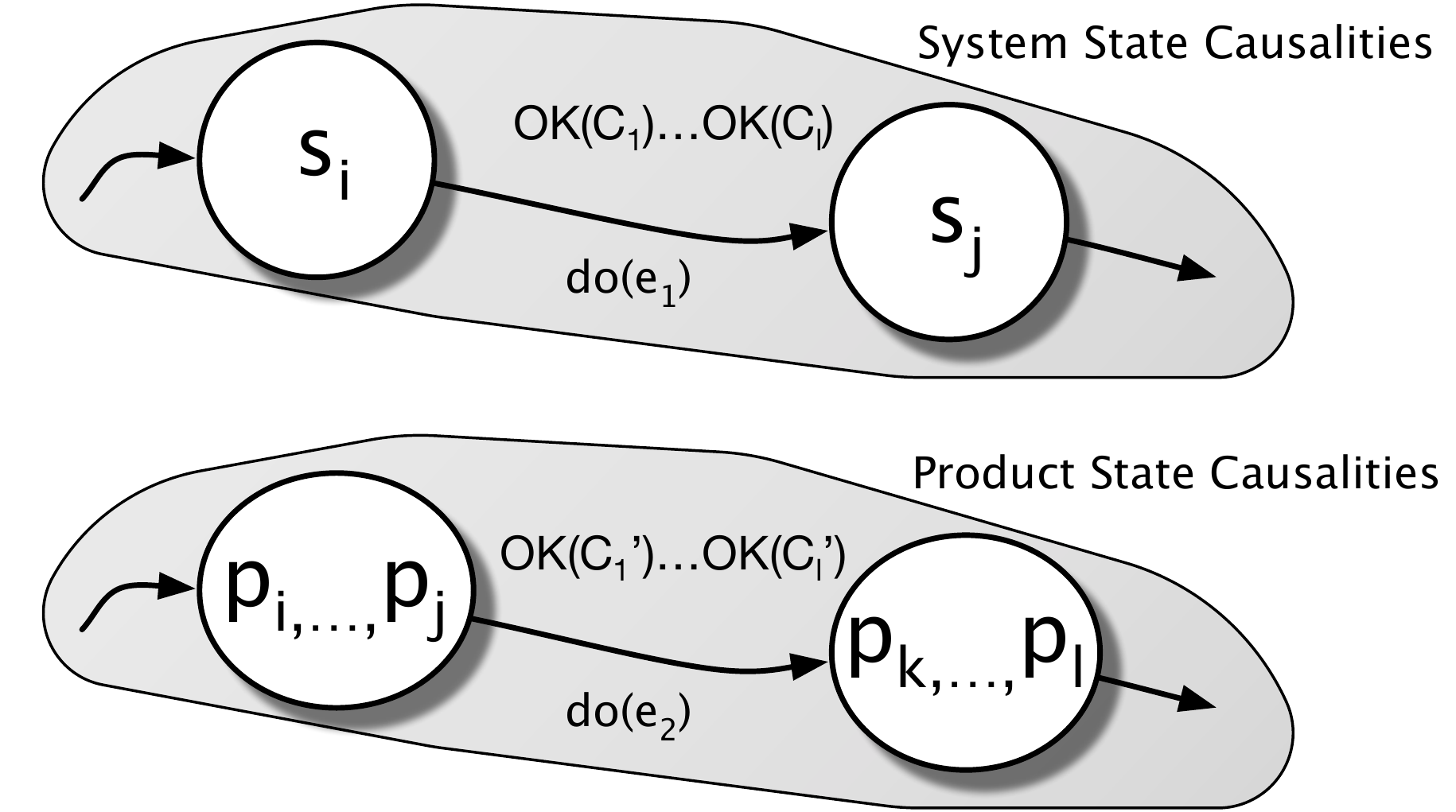}{The main causality concept of the \dt AI reference model AITwin}

\noindent \textit{\bf System State Causalities:} Such causalities model the transition from a system state $s_i \in S$ to state $s_j \in S$. The occurrence of an event $e_1$ triggers, denoted by Pearl's \textbf{do} operator \cite{Pearl:2009:CMR:1642718}, this transition. Please note that also empty events, i.e. immediate transitions without external triggering events, are possible.

A system state $s\in S$ corresponds to a subset of $\mathcal{X}$, with $\mathcal{X}$ being the set of all possible values $\mathbf{x}(t)$. Here we restrict this to convex subsets of $\mathcal{X}$, corresponding to linear inequalities on $\mathbf{x}(t)$ such as $5.5 < x_1 < 12.7\ \& \ 1.2 < x_2 + x_3 < 9.9\ \& \ldots $.

An event $e_1$ is then defined as the crossing of a threshold in $\mathcal{X}$. Such thresholds correspond to equations of the form $\mathbf{f} \cdot \mathbf{x}(t) < \mathbf{c}, \mathbf{f}, \mathbf{x}(t), \mathbf{c} \in \real^n$ which define a halfspace of $X$. Such a system state causality is only valid under the condition that specific components $C_1, \ldots, C_l$ are functioning correctly, e.g. $OK(C_1) \wedge \ldots \wedge OK(C_l)$ is true (see section \ref{extrapolate}). Please note that with these information the diagnosis task from above can be implemented, e.g. consistencies between predictions $\mathbf{x}(t)$ from the predictions API level and causalities predicting system states $s_i$ can be checked as $s_i$ is a subspace of $\mathcal{X}$. 

\smallskip
\noindent \textit{\bf Product State Causalities:} Such causalities model the transition from one set of objects $p_i, \ldots, p_j$ to another set $p_k, \ldots, p_l$. Again, the occurrence of an event $e_2$ triggers this transition and empty events are possible. Events are defined as above. Such a system state causality is only valid under the condition that specific components $C_1', \ldots, C_l'$, see section \ref{extrapolate}, are functioning correctly, e.g. $OK(C_1') \wedge \ldots \wedge OK(C_l')$ is true. 
The resulting API extensions are shown in figure \ref{arch3.pdf}. Please note that with these information the planning task from above can be implemented. 
\bsfigure[htb!]{0.35}{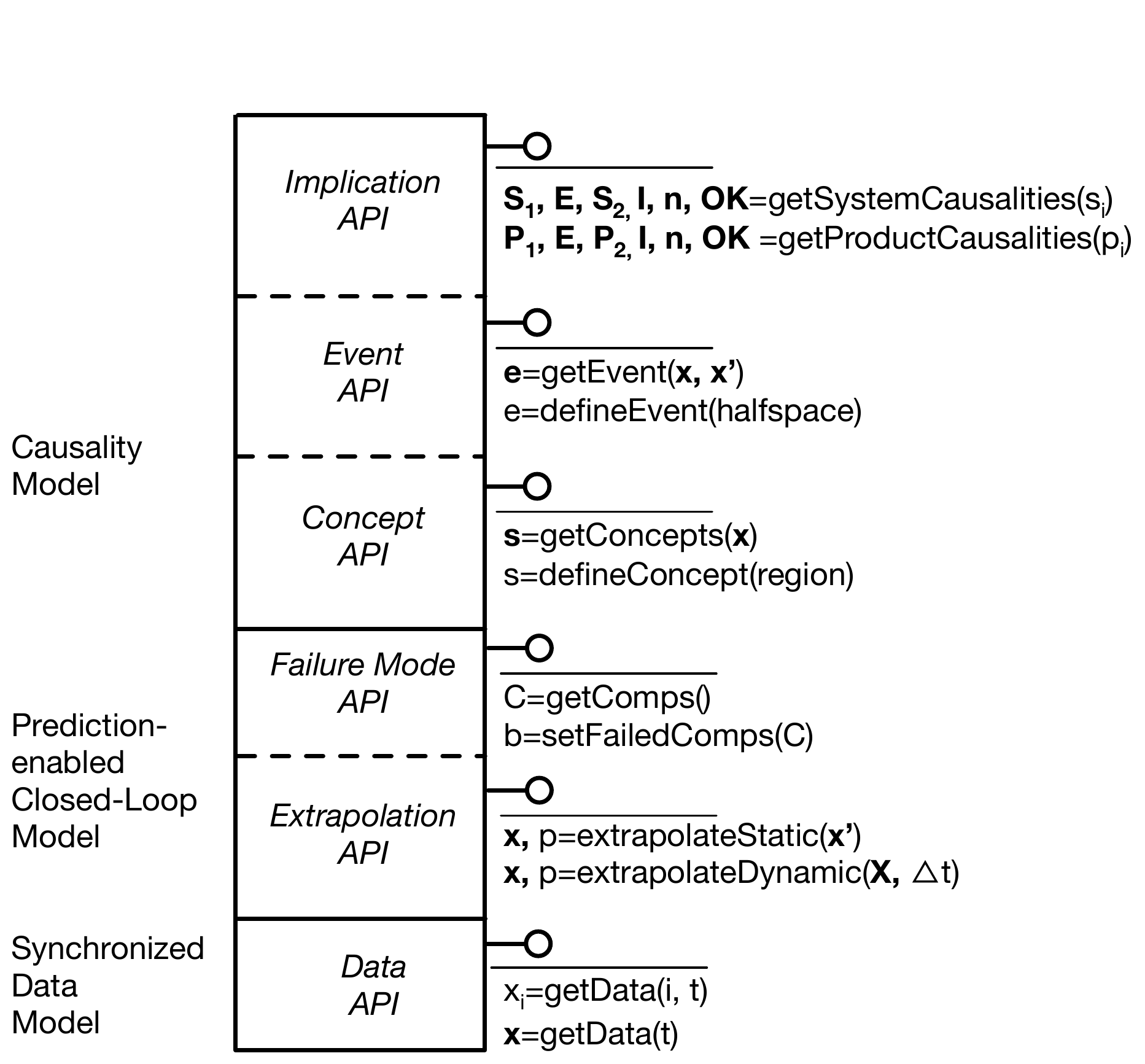}{The complete \dt AI reference model AITwin}

An implication API allows for the handling of causalities in form of transitions or implications: The function 
$\text{getSystemCausalities}(s_i)$ returns all causalities $s_1, e, s_2, I, n, OK$. $s_1$ denotes the starting state, $s_2$ the end state, $e$ is the event, $n$ the name of the causality and $I$ stores additional information such as probabilities, timing etc. $OK$ is the set of $OK$ assumptions $OK = OK(C_1) \wedge \ldots \wedge OK(C_l)$ which are a precondition for the validity of the causality.

The function $\text{getProductCausalities}(p_i)$ returns all causalities $p_1 = (p_1^{1}, \ldots, p_1^{l}), e, p_2 = (p_2^{1}, \ldots, p_2^{k}), I, n, OK$ where $p_1$ denotes the original objects, $e$ is the event, $p_2$ denotes the resulting objects, $n, I, OK$ are defined as above. 

$e=\text{defineEvent(halfspace)}$ defines an event $e$ which corresponds to an halfspace. The parameter \emph{halfspace} is defined using a mathematical inequality in the language MathML \cite{mathml}. The function \emph{getEvent}$(\mathbf{x}, \mathbf{x'})$ returns all events which are triggered when moving from the operation point $\mathbf{x}\in \mathcal{X}$ to $\mathbf{x'}\in \mathcal{X}$. 

$s=\text{defineConcept(region)}$ defines a system state $s$, namely a concept, as a convex region \emph{region}. The parameter \emph{region} is part of in $\mathcal{X}$ using a mathematical inequality in the language MathML \cite{mathml}. The function \emph{getConcepts}$(\mathbf{x})$ returns all system states $s_i$ (aka concepts) with $\mathbf{x} \in s_i$.

%% file: aitwin.tex
\subsection{The AITwin Reference Model} \label{aitwin}
Figure \ref{arch3.pdf} shows the complete \dt AI reference model AITwin. Please note that the focus is on those parts of a \dt which are required by AI algorithms. Further information for domains such as production, automation and logistics must be added. AITwin may be used in parallel on different system granularities, e.g. for machine modules in production plants or for production lines. 

Please note that the three levels of figure \ref{levels.pdf} correspond to the three levels of figure \ref{arch3.pdf} and form a holistic concept: Data $\mathbf{x}$ from the ''Synchronized Data Model''-API is used also in the ''Prediction-enabled Closed-Loop Model''-API. And these data points can also be correlated to system states from that ''Causality Model''-API. And all APIs use the same components for modeling failure modes. 

 Besides those APIs, the \dt is based on three state spaces which must be part of the \dt definition: $\mathcal{X}$ denotes all possible signal values, $P$ denotes all possible object/product configurations and $COMPS$ is the set of components which may fail. Furthermore, two derived state spaces exist, which normally are defined via API methods during the operation phase: $S$ is a set of system states where $s\in S$ is a subset of $\mathcal{X}$ and $E$ is the set of all events where $e\in E$ is a halfspace of $\mathcal{X}$.
 
 Please note that the APIs here do not cover the way how the information are added to the DigitalTwin. 

%% file: empiric2.tex
\section{Evaluation} \label{empirical}

Two ways exists to evaluate such a new API. In section \ref{exprocess}, AITwin is used by several AI algorithms for one running example. In section \ref{projects}, previous research projects in the field of AI in production are analyzed whether AITwin would have been suitable.

\subsection{Applying AITwin to a Running Example} \label{exprocess}

In a next step, the ''synchronized data'' and especially the ''extrapolation'' APIs have been used to detect anomalies. Anomalies can be inserted via the ''failure mode'' API. 

For this, first data is extracted via the ''data'' API and used to train a model of the normal behavior via unsupervised learning. Later, non-normal data is fed to the models and the models are supposed to detect these anomalies. As learning algorithms, Support Vector Machines (SVM) and Long-Short-Term-Memory (LSTM) neural network are used. Table \ref{tab:results} shows the results for different systems and different faults. Both Area-Under-Curve (AUC) and also F1 measures show that anomalies are detected with a high quality.

\begin{table}[t]
    \begin{scriptsize}
        \centering
        \begin{tabular}{|l|l|c|c|c|c|}
            \hline
            \rowcolor{Gray}
            Config. & Fault & \multicolumn{2}{c|}{LSTM Autoen.} & \multicolumn{2}{c|}{SVM} \\
            \rowcolor{Gray}
            & & AUC & F1 & AUC & F1 \\
            \hline
            a\_tank & stable & 1.000 & 1.000 & 1.000 & 1.000 \\
            a\_tank & pumpSlow & 0.500 & 0.406 & 0.997 & 0.997\\
            a\_tank & tank1\_leak & 0.500 & 0.402 & 0.994 & 0.994\\
            a\_tank & valve0Block & 0.989 & 0.977 & 0.999 & 0.999\\
            a\_tank & valve1Block & 0.984 & 0.950 & 0.990 & 0.970\\
            a\_tank & valve1Stuck & 0.989 & 0.977 & 0.500 & 0.666\\
            4\_tanks & stable & 1.000 & 1.000 & 1.000 & 1.000 \\
            4\_tanks & pipe4\_jam & 0.996 & 0.996 & 0.992 & 0.992 \\
            4\_tanks & tank2\_leak & 0.960 & 0.959 & 0.992 & 0.992 \\
            4\_tanks & valve3\_jam & 0.996 & 0.996 & 0.983 & 0.983 \\
            4\_tanks & valve6\_jam & 0.996 & 0.996 & 0.701 & 0.770 \\
            3\_tanks & stable & 1.000 & 1.000 & 1.000 & 1.000\\
            3\_tanks & pumpFast & 0.915 & 0.833 & 0.727 & 0.588\\
            3\_tanks & pumpSlow & 0.993 & 0.990 & 0.929 & 0.562 \\
            3\_tanks & tank1Leak & 0.994 & 0.992 & 0.517 & 0.554 \\
            3\_tanks & tank2Leak & 0.994 & 0.992 & 0.959 & 0.960\\
            3\_tanks & valve2Closed & 0.988 & 0.978 & 0.985 & 0.895 \\
            3\_tanks & valve3Closed & 0.986 & 0.975 & 0.505 & 0.057 \\
            \hline
        \end{tabular}
        \caption{Results}
        \label{tab:results}
    \end{scriptsize}
\end{table}

\smallskip
In a second step, the AITwin is applied to symbolic AI tasks. For this, concepts are extracted from the DigitalTwin via the ''concept'' API and causalities via the ''implication'' API. This information can be used to create a model for Consistency-Based Diagnoses (see section \ref{causalities}). Failures can be added via the ''failure'' API. Table shows the results, most failures can be identified correctly.

We also used a quantitative simulation of the Tennessee Eastman process. The implementation of Downs et al. \cite{downs1993plant} contains 20 different injected faults (process disturbances). However, the instrumentation of the simulated process is such that not all faults will be identified exactly. Table \ref{tbl:results_tesim} shows the results for six experiments. The injected faults for IDV 16 through 20 have an insufficient description for validating results. The other faults not described contain fault modes that are not associated with single components and thus cannot be evaluated. This leaves us with the experiments shown in Table \ref{tbl:results_tesim}.
\begin{table}[ht]
	\centering
	\label{tbl:results_tesim}
	\begin{tabular}{c|c|l}
		IDV & Fault isolated & Injected Fault   \\ \cline{1-3}
		$1$   &  not ok             & Feed ratio changed    \\ 
		$6$  &  ok             & Pipe A feed loss    \\ 
		$8$  & ok              &  Feed ratio changed   \\ 
		$13$   & not ok             &  Reactor kinetics fault   \\ 
		$14$  & ok               &  Reactor cooling fault    \\ 
		$15$  & ok             & Condenser cooling fault  \\ 
	\end{tabular}
	\caption{Experimental results with the Tennessee Eastman process}
\end{table}

An x indicates that the faulty component was found as part of the minimal cardinality diagnosis and a - shows that the faulty component was not found. We were able to find all faults that had an identifiable component fault as the cause of the process disturbance. The change of input ratios can only be detected indirectly, since no observations are available at the inputs. 

These two result tables show clearly that the AITWin concept can be used for a wide range of AI applications---from data-based machine learning problems to symbolic AI diagnosis problems.

\subsection{AITwin and Research Projects} \label{projects}

In the following, some projects and publications from the field of ''AI for Production'' are listed and analyzed with regard to the AITwin. For this we split those references into the field of condition-monitoring, predictive maintenance, optimization and the field diagnosis and planning.

\vspace*{-5px}
\subsection{Condition Monitoring, Predictive Maintenance, Optimization}
There is a huge number of  projects and research works in the field of machine learning regarding anomaly detection, condition monitoring, predictive maintenance and optimization and the following review can only cover a small part of this work. Starting with the project SMARTPas \cite{thowl1}, which deals with the prediction of beverage characteristics, the data \& prediction API can be used to predict new data points for control and optimization. The projects IMPROVE \cite{thowl2} and PrognoseBrain \cite{thowl4}, dealing with condition-monitoring and anomaly detection in production systems with ML, could have used the data \& prediction API asw well. In terms of research works e.g. Li \cite{li2019prediction}, who predicts the surface roughness in extrusion-based additive manufacturing could have used the proposed DigitalTwin. Eiteneuer \cite{eitenheuer2018jdgsw}, performing an anomal detection with LSTMs, can use the data \& prediction API as well. Other examples in which the DigitalTwin could have been used can be found in \cite{wu2017comparative}, giving a comparative survey with ML in manufacturing, in \cite{cavalcante2019supervised} using ML in digital manufacturing or in \cite{caggiano2019machine} who does an on-line defect recognition. Within all these projects and research work, instead of having several incompatible data models, possibly resulting in not comparable results, one could have used a single DigitalTwin and the data \& prediction API.

\vspace*{-5px}
\subsection{Diagnosis and Planning} \label{symbolicai}
This section reviews several typical projects and research work in the field of symbolic AI, mainly covering diagnosis and planning tasks. Especially Diagnosis algorithms can, if consistency-based, get all necessary information from the AITwin reference models. It needs to be kept in mind that other diagnosis methods can be added, but within the AITwin concept consistency-based diagnosis is especially suited for CPS---please refer to \cite{diedrich2019jfgsy} for details. Regarding the projects, e.g. EfA \cite{boettcher2013design}, which deals with the development of intelligent assistance systems, can use the causality API for the planning of automation systems. The project IMPROVE \cite{thowl2}, in which innovative modelling approaches for production systems are developed, can use the causality API for the detection of system causalities. In terms of recent publications, e.g. \cite{Rogalla:2017}, aiming to do an automated process planning for CPS and \cite{peng201919b} doing non-convex hull based anomaly detection in cyber-physical prodution systems could use the AITWin. Other examples are \cite{ademujimi2017review} who gives a detailed review of ML for diagnoses in production or \cite{zhu2019improved} who analysis gas wells. In every work the causality API of the AITwin can be used. \newline
Regarding planning tasks, projects and research work can also derive most of its needed information from the AITwin. So far, descriptions of machines must be implemented individually but can refer to process step descriptions as capability models. Furthermore, the computed overall production process is also stored individually, but could be added in a straight forward manner. The bottom-line is that the analyzed consistency-based diagnosis activities for CPS can be implemented with the AITwin model. The planning activities can to a large extent be implemented with the AITwin model, some proprietary information must be added.

%% file: contribution.tex
\section{Conclusion} \label{contribution}
In section \ref{intro}, three research questions have been identified as key challenges for the definition of a common DigitalTwin reference model, called AITwin. For this, section \ref{section3} analyzes conceptual requirements of CPS to timing models, section \ref{extrapolate} analyzes conceptual requirements of CPS to machine learning models and section \ref{causalities} analyzes conceptual requirements of CPS to symbolic AI models. Based on these requirements, concepts and APIs are derived in section \ref{aitwin}. This AITwin answers the three research questions. 
 
Section \ref{empirical} uses a running example and projects to empirically verify the AITwin reference model leading to the point that most AI and ML activities can be implemented using it. The authors do not claim that the reference models is the ultimate definition of a common AI DigitalTwin. But it is to a significant extent able to meet the requirements and may serve as a crystallization point for further research. \

Assuming that both DigitalTwins and AI/ML methods are crucial elements of future CPS architectures, which is a rather safe bet, both worlds need to be harmonized. The AITwin reference model can prevent unnecessary redundant modeling efforts and allows for a reuse and sharing of computed results. Finally, this can be a starting point for a standardization of an AI-enabled DigitalTwin for CPS. 

\noindent \textit{\bf Acknowledgements:} 
This work was partly developed within the Fraunhofer Cluster of Excellence "Cognitive Internet Technologies".